# Detection and Segmentation of Cosmic Objects Based on Adaptive Thresholding and Back Propagation Neural Network

Samia Sultana[1*] Shyla Afroge[2]

[1]Computer Science and Engineering
Rajshahi University of Engineering and Technology, Rajshahi-6204, Bangladesh
[2]Lecturer, Computer Science and Engineering
Rajshahi University of Engineering and Technology, Rajshahi-6204, Bangladesh
[*]samia.cse.ruet@gmail.com

*Abstract*— Astronomical images provide information about the great variety of cosmic objects in the Universe. Due to the large volumes of data, the presence of innumerable bright point sources as well as noise within the frame and the spatial gap between objects and satellite cameras, it is a challenging task to classify and detect the celestial objects. We propose an Adaptive Thresholding Method (ATM) based segmentation and Back Propagation Neural Network (BPNN) based cosmic object detection including a well-structured series of pre-processing steps designed to enhance segmentation and detection.

## I. INTRODUCTION

We have proposed sequential order of pre-processing steps before segmentation and cosmic object detection. Segmentation based on Adaptive thresholding Method and back propagation neural network algorithm for learning and recognizing celestial objects. Astronomical image processing system faces so many obstacles during segmentation and detection. To overcome these obstacles, multiple pre-processing steps are performed on the actual image prior to segmenting the desired object(s).

## II. DATA

Input of this work is images of celestial objects. The standard data format used in astronomy is FITS that stands for 'Flexible Image Transport System'. The Charge Coupled Device (CCD) images obtained from celestial observations are usually stored in FITS format. This input image is in FITS format. This format is endorsed by NASA and the International Astronomical Union (IAU). This is used for transport, analysis, and archival storage of scientific data sets.

Two data is used here for experiment. One data is Eagle Nebula and Another is Comet Ison. Eagle Nebula was captured by "Faulkes Telescope North". It is a huge data of 1026x1024 dimensions. The Eagle Nebula is a cluster of stars discovered by Jean-Philippe in 1745-46[1]. The input data 2 is Comet ISON which is an image of Comet. This input image contains 1074x1074 dimensions, which mean the data is 1074 pixels width and 1074 pixels height.

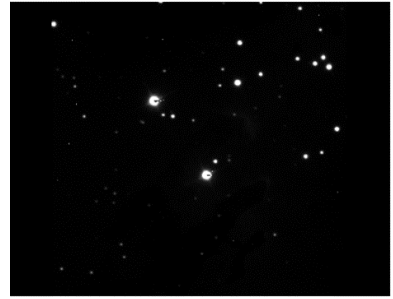

Figure 1: Eagle Nebula (**Source:** Las Cumbres Obsevatory Global Telescope Network) [2]

## III. METHODOLOGY

The methodology contains different types of steps to segment astronomical images. The steps are as follows-

### A. Image Enhancement

Enhancing an image provides better contrast and a more detailed image as compared to non-enhanced image. Here image is enhanced by Log Transformation.[3] The log transformations can be defined by this formula:

$$s = c\,log(r + 1)$$

Where s and r are the pixel values of the output and the input image, c is a constant.

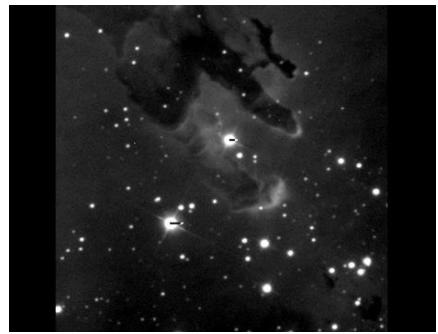

Figure 2: Log Transformation

### B. Morphological Adjustments

Astronomical images usually contain numerous bright point sources (stars, distant galaxies etc.). These point sources are to be primarily removed for the proper segmentation in the final stage. The removal of such sudden change in intensities prevents the evolving level set contour, from getting stuck at the local regions. Local peak search using a matched filter or a cleaning process [5] are usually used to remove the unresolved point sources. But multiple pass through the filter, which is required during these processes, would diffuse the image and may result in the break-up of the components of the extended sources [4][5]. Erosion is a morphological enhancement process [6], would serve the purpose without affecting the shape and structure of the celestial objects to an extent. We have removed the peak intensities and bright point sources using erosion.

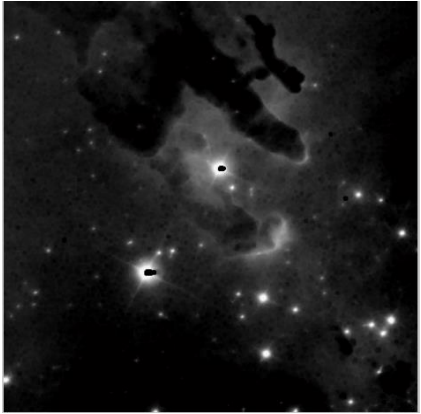

Figure 3: Erosion

*C. De Noising*

A **Gaussian filter** is a filter whose impulse response is a gaussian function. Gaussian filters have the properties of having no overshoot to a step function input while minimizing the rise and fall time. This behaviour is closely connected to the fact that the Gaussian filter has the minimum possible group delay. It is considered the ideal time delay filter, just as the sinC is the ideal frequency domain filter. [7]. In one dimension, the Gaussian function is:

$$G(x) = \frac{1}{\sqrt{2\pi\sigma^2}} e^{-\frac{x^2}{2\sigma^2}}$$

Where, σ is the standard deviation of the distribution.

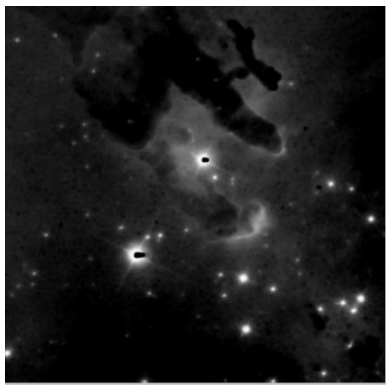

**Figure 5**: De noising by Gaussian Filtering

*D. Segmentation*

Here we have used the local Adaptive threshold method to segment the astronomical image. The Local Adaptive Threshold is used in uneven lighting conditions when you need to segment a lighter foreground object from its background. In many lighting situations shadows or dimming of light cause thresholding problems as traditional thresholding considers the entire image brightness. Adaptive Thresholding will perform binary thresholding (i.e., it creates a black and white image) by analysing each pixel with respect to its local neighbourhood. This localization allows each pixel to be considered in a more adaptive environment. It will perform binary Thresholding by analyzing each pixel with respect to its local neighborhood and calculating the midrange of current pixel. We focus on the binarization of image documents using local adaptive thresholding technique, because in Global thresholding methods such as one proposed by Otsu [8] try to find a single threshold value for the whole document. But it may lose huge amount of data. That's why we have used local Adaptive threshold method.

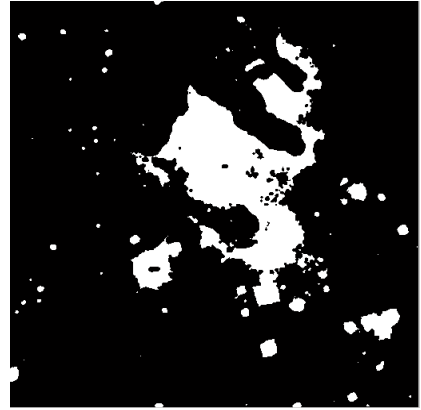

Figure 4: Segmentation by adaptive Thresholding

*E. Object Detection by BPNN*

It is a supervised learning technique with multi-layer perceptron. First proposed by Paul Werbos in the 1970's and rediscovered by Rumelhart and McClelland in 1986[9].It is a training procedure which allows Multilayer feed-forward Neural Network to be trained.

The layer that is not input or output is called hidden layer. Objective of this neural network method is to classify successfully the non-linearly separable data.

There are three layers of this network:

- **Input Layer:** Introduces input value into the network, No activation function or other processing.
- **Hidden Layer:** Perform classification of features, It works like a black box and feed input to output layer.
- **Output Layer:** Functionally just like the hidden layers, Outputs are passed on to the world.

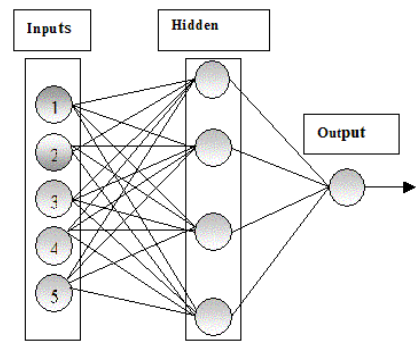

**Figure 6:** architecture of a back propagation neural network

To detect an object, it is necessary to match the pattern. For pattern matching 10 hidden layers and 7 input nodes with some randomize weighted values. We have two sessions; one is a learning session, and another is testing session. During these two sessions 3 types of samples are taken as follows:
1. **Training Samples:** These are presented to the network during training, and the network is adjusted according to its error.
2. **Validation Samples:** These are used to measure network generalization, and to halt training when generalization stops improving.
3. **Testing Sample:** These have no effect on training and so provide an independent measure of network performance during and after training.

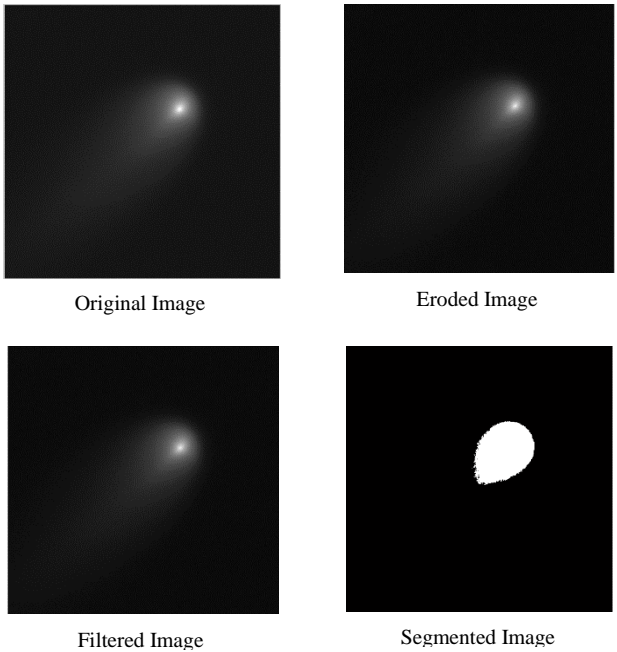

Original Image | Eroded Image

Filtered Image | Segmented Image

**Figure 6:** Step by step output image of Comet ISON detection

IV. RESULT ANALYSIS

A. *Adaptive Threshold Values of Output Image*

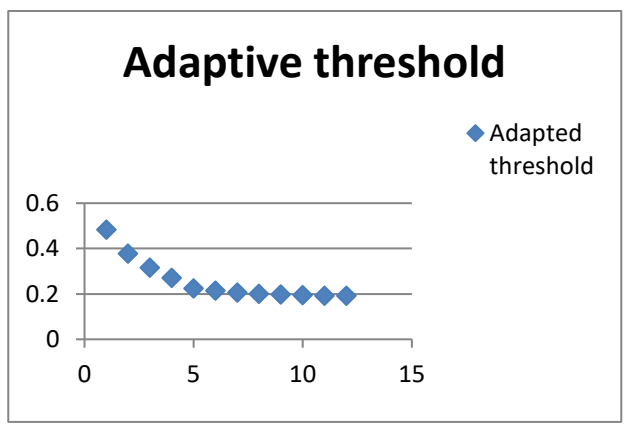

**Figure 7:** Graph of Iteration number vs Adaptive threshold values (Data 1)

TABLE 1
THRESHOLD VALUES AND ERRORS FOR DATA 1

| Iteration Number | Threshold T | Error[abs(T-T0)] |
|---|---|---|
| 1 | 0.3782 | 0.1061 |
| 2 | 0.3156 | 0.0626 |
| 3 | 0.2711 | 0.0446 |
| 4 | 0.2430 | 0.0281 |
| 5 | 0.2240 | 0.0190 |
| 6 | 0.2145 | 0.0095 |
| 7 | 0.2069 | 0.0076 |
| 8 | 0.2014 | 0.0055 |
| 9 | 0.1985 | 0.0029 |
| 10 | 0.1955 | 0.0030 |
| 11 | 0.1926 | 0.0029 |
| 12 | 0.1926 | 0 |

TABLE 2
THRESHOLD ACCURACY OF ADAPTIVE SEGMENTATION FOR DATA 1

| Initial Threshold $T_0$ | Adaptive Threshold T | MSE | PSNR | Error Rate % | Accuracy |
|---|---|---|---|---|---|
| 0.4843 | 0.1926 | 0.1146 | 57.5395 | 11.46% | 89.63% |

TABLE 3
ACCURACY OF BPNN BASED DETECTION FOR DATA 1

| Statistics | Training Set | Test Data |
|---|---|---|
| Accuracy | 66.2% | 58.3% |
| MSE | 0.332 | 0.627 |
| Data 1 Error | 32.17% | 61.02% |

TABLE 4
THRESHOLD ACCURACY OF ADAPTIVE SEGMENTATION FOR DATA 2

| Initial Threshold T0 | Adaptive Threshold T | MSE | PSNR | Error Rate % | Accuracy |
|---|---|---|---|---|---|
| 0.4824 | 0.1810 | 0.0580 | 60.4939 | 5.80% | 94.22% |

TABLE 5
ACCURACY OF BPNN BASED DETECTION FOR DATA 2

| Statistics | Training Set | Test Data |
|---|---|---|
| Accuracy | 65.4% | 57.93% |
| MSE | 0.304 | 0.631 |
| Data 2 Error | 31.27% | 42.89% |

## V.    CONCLUSIONS

It is always challenging to process the astronomical images. The original images from the telescope archive are rarely available as high confidentiality is maintained, so lots of data are missing in our input images. Color images are available, but the images directly captured by telescopes are not available. Some parts of the Celestial objects are often very faint due to the presence of bright point sources. The images have low contrast due to long distances and disturbances. The celestial objects have lack of clear-cut boundaries. As experimented data size is huge, e.g., one is 1026x1024 dimensions and another is 1074x 1074 dimensions, so the detection accuracy is low. The detection technique takes long time to execute (more than 1 hour). The performance of detection is not so good and accuracy is very low. The accuracy is varying with different data.